\documentclass[a4paper]{article}

\usepackage{INTERSPEECH_v2}
\DeclareMathOperator*{\argmax}{arg\,max}

\title{Comparison of Decoding Strategies for CTC Acoustic Models}
\name{Thomas Zenkel$^{1,2}$, Ramon Sanabria$^1$, Florian Metze$^1$, Jan Niehues$^2$,\\
Matthias Sperber$^2$, Sebastian St\"uker$^2$ and Alex Waibel$^{1,2}$}
\address{
  $^1$Carnegie Mellon University; Pittsburgh, PA; U.S.A.\\
  $^2$Karlsruhe Institute of Technology; Karlsruhe, Germany
}
\email{
  \{thomas.zenkel, jan.niehues, matthias.sperber, sebastian.stueker\}@kit.edu \\ 
  \{ramons, fmetze, ahw\}@cs.cmu.edu\\
}

\begin{document}

\maketitle
\begin{abstract}
   Connectionist Temporal Classification has recently attracted a lot of interest as it offers an elegant approach to building acoustic models (AMs) for speech recognition. The CTC loss function maps an input sequence of observable feature vectors to an output sequence of symbols. Output symbols are conditionally independent of each other under CTC loss, so a language model (LM) can be incorporated conveniently during decoding, retaining the traditional separation of acoustic and linguistic components in ASR.
   
   For fixed vocabularies, Weighted Finite State Transducers provide a strong baseline for efficient integration of CTC AMs with n-gram LMs. Character-based neural LMs provide a straight forward solution for open vocabulary speech recognition and all-neural models, and can be decoded with beam search. Finally, sequence-to-sequence models can be used to translate a sequence of individual sounds into a word string.
   
   We compare the performance of these three approaches, and analyze their error patterns, which provides insightful guidance for future research and development in this important area.
\end{abstract}

\section{Introduction} 

Traditionally, Acoustic Models (AMs) of an Automatic Speech Recognition system followed a generative approach based on HMMs \cite{rabiner1986introduction} where the emission probabilities of each state were modeled with a Gaussian Mixture Model. Since the AM works with phonemes as a target, during decoding the information of the AM had to be combined with a pronunciation lexicon, which maps sequences of phonemes to words, and a word based LM \cite{soltau2001one}.

More recent work has been focused on solutions which come close to end-to-end systems. Connectionist Temporal Classification (CTC) acoustic models \cite{graves2006connectionist} can directly model the mapping between speech features and symbols without having to rely on an alignment between the audio sequence and the symbol sequence. However, the CTC objective function requires that its output symbols are conditional independent of each other. While this assumption is essential to learn a mapping between the speech features and the output sequence, it also entails to add linguistic information during decoding.

Other end-to-end approaches that are inspired by recent developments in machine learning system such as \cite{bahdanau2014neural} are \cite{bahdanau2016end, chan2015listen}. By attending to different frames for each output symbol attention based speech recognition systems are able to map speech features to an output sequence. 

This approach has no need to assume conditional independence between its output, and therefore is theoretically able to jointly learn acoustic and linguistic models implicitly.

While ``traditional'' from a strictly end-to-end point of view, the separation of acoustic model and language model allows for domain independence and adaptation or re-use of speech recognition components. 
In this work we therefore investigate different decoding strategies for CTC acoustic models, and provide a comparison of their individual characteristics using the same acoustic model. We provide a performance comparison, and analyze the differences in the output of the model.


We compare the following approaches:

\begin{itemize}
  \item Greedy Search without linguistic information
  \item WFST search with word language model \cite{miao2015eesen,sak2015learning}
  \item Beam search using character RNN language model \cite{maas2015lexicon, hwang2016character}
  \item Sequence to Sequence approach using neural machine translation
\end{itemize}

For each of these categories we implement a search algorithm and evaluate it on the Switchboard Task, thereby providing a fair comparison of the most promising decoding approaches. 

\section{Related work} 
The simplest decoding algorithm is to pick the most likely character at each frame. This is commonly used to provide Character Error Rates (CER) during training of the acoustic model and can also be used to calculate Word Error Rates (WER), given that the acoustic model has a notion of word boundaries. Word boundaries can be modeled with a space symbol or by capitalizing the first letter of each word \cite{zweig2016advances}. While decoding CTC acoustic models without adding external linguistic information works well, a vast amount of training data should be used to get competitive results \cite{soltau2016neural}.

A traditional approach to perform decoding over CTC is to add linguistic information on the word level. Early work did this with an ordinary beam search, that means by performing a breadth first search in the time dimension and keeping a fixed number of partial transcriptions at every time step. For including linguistic information when adding a new character to a transcription, word based LMs were preprocessed \cite{graves2014towards, hannun2014first}.

Weighted Finite State Transducers (WFST) present a generalized word based approach  \cite{miao2015eesen, mendis2016parallelizing}. WFST provide a convenient framework to combine a word based n-gram model and a lexicon which maps a sequence of symbols to a word into a single search graph. While this allows to process both character and phonemes as the output of the acoustic model, it can only generate sequences of words from a fixed vocabulary.

Due to recent developments of character based LMs \cite{foerster2016intelligible, krause2016multiplicative}, it is also a competitive option to directly add character level linguistic information during the beam search. Currently, one of the most promising approaches is to use a character based RNN and query it when a new character is added to the transcription \cite{maas2015lexicon, hwang2016character}. With its theoretically infinite context a character RNN can encourage the transcription to be linguistically correct while adding its information as soon as possible.

The last approach presented is to treat the decoding problem as a general sequence to sequence task. For each frame the acoustic model outputs a probability distribution over all labels. This information can be processed by another CTC model \cite{zweig2016advances} or by an attention based system \cite{bahdanau2014neural} to produce a more linguistically reasonable transcription. Recent approaches combine a CTC model with an attention based mechanism and are able to train this model jointly \cite{kim2016joint}.

\section{Acoustic Model} 
The AM of our system is composed by multiple RNN layers followed by a soft-max layer. RNN layers, which are composed by bidirectional LSTM units~\cite{hochreiter1997long}, provide the ability to learn complex, long term dependencies. A sequence of multiple speech features forms the input of our model. For each input the AM outputs a probability distribution over its target alphabet. The whole model is jointly trained under the CTC loss function~\cite{graves2006connectionist}.
 
More formally, let us define a sequence of $n$-dimensional acoustic features $\bm{X} = (\bm{x}_1,\dots, \bm{x}_T)$ of length $T$ as the input of our model and $L$ as the set of labels of our alphabet. These labels can be either characters or phonemes. 
We augment $L$ with a special blank symbol $\emptyset$ and define $L' = L \cup \emptyset$.

Let $\bm{z} = (z_1, ..,z_U) \in L^{U}$ be an output sequence of length $U \leq T$, which can be seen as the transcription of an input sequence. To define the CTC loss function we additionally need a many to one mapping $\mathcal{B}$ that maps a path $\bm{p} = (p_1, \dots p_T) \in L'^T$ of the CTC model to an output sequence $\bm{z}$. This mapping is also referred as the squash function, as it removes all blank symbols of the path and squashes multiple repeated characters into a single one (e.g. $\mathcal{B}(AA \emptyset AAA BB) = AAB$). Note that we do not squash characters that are separated by the blank symbol as this still allows us to create repeated characters in the transcription. Let us define the probability of a path as
\begin{equation}
P(\bm{p}|\bm{X}) = \prod_{t=1}^T y_k^t
\end{equation}
where $y_k^t$ is the probability of observing the label $k$ at time~$t$. To calculate the probability of an output sequence $\bm{z}$ we sum over all possible paths:
\begin{equation}
P(\bm{z}|\bm{X}) = \sum_{\bm{p}\in \textit{B}^{-1}(\bm{z})}P(\bm{p}|\bm{X})
\end{equation}
To perform the sum over all path we will use a technique inspired by the traditional dynamic programming method used in HMMs, the forward-backward algorithm~\cite{rabiner1986introduction}. We additionally force the appearance of blank symbols in our paths by augmenting the sequence of output labels during training with a blank symbol between each of the labels of $\bm{z}$ as well as at the beginning and the end of the sequence. 

Given a sequence of speech features $\bm{X} = (\bm{x_1}, \dots, \bm{x_T}$), we can now calculate the probability distribution over the augmented label set $L'$ for each frame. In the remainder of the paper let $P_{AM}^t(k|\bm{X})$ denote the probability to encounter label $k \in L'$ at time step $t$ given the speech features $\bm{X}$. The decoding strategies of the subsequent chapter will process this information in different ways to create a linguistically reasonable transcription.

\section{Decoding Strategies}
In this section we describe different approaches on how to generate a transcription given the static sequence of probabilities generated by the acoustic model.

\subsection{Greedy Search}
To create a transcription without adding any linguistic information we use the decoding procedure of \cite{graves2006connectionist} and greedily search  the best path $p \in L'^{T}$:
\begin{equation}
\argmax_{\bm{p}} \prod_{t=1}^T P_{AM}^t(p_t|\bm{X})
\end{equation}
The mapping of the path to a transcription $z$ is straight forward and works by applying the squash function: $z = \mathcal{B}(p)$. For character based CTC acoustic models this procedure can already provide useful transcriptions.

\subsection{Weighted Finite State Transducer}\label{wfstdef}
To improve over the simple greedy search, the Weighted Finite State Transducer (WFST) approach adds linguistic information at the word level. First of all we preprocess the probability sequence with the prior probability of each unit of the augmented label set $L'$. 
\begin{equation}
p(\bm{X}|k) \propto P(k|\bm{X}) / P(k)
\end{equation}
This does not have a proper theoretical motivation since the result is only proportional to a probability distribution.  However, by dividing through the prior probability units which are more likely to appear at a particular position than their average will get a high value. 

The search graph of the WFST is composed of three individual components:
\begin{itemize}
	\item A token WFST maps a sequence of units in $L'$ to a single unit in $L$ by applying the squash function $\mathcal{B}$
  \item A lexicon WFST maps sequences of units in $L$ to words
  \item A grammar WFST encodes the permissible word sequences and can be created given a word based n-Gram language model
\end{itemize}
The search graph is used to find the most probable word sequence. Note that the lexicon of the WFST allows us to deal with character as well as phoneme based acoustic models.

\subsection{Beam Search with Char RNN}
In contrast to the WFST based approach we can directly apply the probabilities at the character level with this procedure. For now assume that the alphabet of the character based LM is equal to $L$. We want to find a transcription which has a high probability based on the acoustic as well as the language model. Since we have to sum over all possible paths $p$ for a transcription $z$ and want to add the LM information as early as possible, our goal is to solve the following equation:
\begin{equation}
\argmax_{\bm{z}} \sum_{\mathcal{B}^{-1}(z) = p} \prod_{t=1}^{T}
y_{p_t}^{t} \cdot P'_{LM}(p_t|\mathcal{B}(\bm{p}_{1:t-1}))
\end{equation}
Note that we cannot estimate a useful probability for the blank label $\emptyset$ with the language model, so we set $P'_{LM}(\emptyset | p) = 1 \forall p \in \mathcal{P}(L') $. To not favor a sequence of blank symbols, we apply an insertion bonus $b \in \mathbb{R}$ for every $p_t \neq \emptyset$. This yields the following equation:
\begin{equation}
P_{LM}'(k|\bm{p})= 
\begin{cases}
    P_{LM}(k|\bm{p}) \cdot b,& \text{if } k \neq \emptyset\\
    1,              & \text{if } k = \emptyset
\end{cases}
\end{equation}
where $P_{LM}(k|\bm{p})$ is provided by the character LM.
As it is infeasible to calculate an exact solution to equation 6, we apply a beam search similar to \cite{hwang2016character}\footnote{Code is included within EESEN: https://github.com/srvk/eesen}.

For AMs which do not use spaces nor have another notion of word boundaries, it is possible to add this information based only on the character LM. This can be achieved by adding a copy of each transcription appended by the space symbol at each time step. This works surprisingly well, since spaces at inappropriate position will get a low LM probability. To the best of our knowledge this is a novel approach and can easily be extended to a larger number of characters, for example to punctuation marks.

While this approach is only able to deal with character based acoustic models, it can create arbitrary, open vocabulary transcriptions.

\subsection{Attention Based Sequence to Sequence Model}
The attention based approach is an example for a sequence to sequence model. We apply greedy search to the information provided by the acoustic model and get a sequence of units $z \in \mathcal{P}(L)$. This sequence provides the input to the attention based system. As in common neural machine translation models the input gets transformed into a sequence of words.

Therefore the system first encodes the character sequence using a RNN-based encoder, creating a sequence of hidden representations $h = (h_1, \dots, h_T)$
of length T. During decoding we calculate an attention vector $a = (a_1, \dots, a_T)$ with $\sum_{t=1}^{T} a_t = 1$ for each output word based on the current hidden state of the decoder. With the hidden representation and the attention vector we can now calculate the context vector c:
\begin{equation}
c = \sum_{t=1}^{T}a_t \cdot h_t
\end{equation}
The decoder uses the context vector to create a probability distribution over the vocabulary of output words. During decoding beam search is used to find the most probable word sequence given the input sequence of characters. By transforming the input to a word sequence the attention model is able to add linguistic information and create an improved transcription.

\section{Training}
This section describes the training process of the acoustic model and the linguistically motivated models used in the different decoding approaches.

\subsection{Acoustic Model}\label{acoustic_model}
We use the Switchboard data set (LDC97S62) to train the AM. This data set consists of 2,400 two-sided telephone conversations with a duration of about 300 hours. It is composed of over 500 speakers with different US accents talking about 50 randomly picked topics. We pick 4000 utterances as our validation set for hyper parameter tunning.
Our target labels are either phonemes or characters.

We also augment the training set to get a more generalized model using two techniques. First, by reducing the frame rate, applying a sub sampling and finally adding an offset we augment the number of training samples. Second, we augment our training set by a factor of 10 applying slight changes to the speed, pitch and tempo of the audio files. The model consist of five bidirectional LSTM layers with 320 units in each direction. It is trained using EESEN~\cite{miao2016empirical}.

\subsection{Weighted Finite State Transducer}

As stated in section \ref{wfstdef} our WFST implementation is composed by three individual components. These components are implemented using Kaldi's~\cite{povey2011kaldi} FST tools. 
We determine the weights of the lexicon WFST by using a lexicon that maps each word to a sequence of CTC labels. The grammar WFST is modeled by using the probabilities of a trigram and 4-gram language model smoothed with Kneser-Ney~\cite{chen1996empirical} discounting. We create the language model based on Fisher transcripts and the transcripts of the acoustic training data using SRILM \cite{stolcke2002srilm}.

\subsection{Character Language Model}
We train the Character LM with Fisher transcripts (LDC2004T19, LDC2005T19) and the transcripts of the acoustic training data (LDC97S62). Validation is done on the transcription of the acoustic validation data. These transcriptions are cleaned by removing punctuation marks and duplicate utterances. This results in a training text of about 23 million words and 112 million characters. The alphabet of the character LM consists of 28 characters, a start and end of sentence symbol, a space symbol and a symbol representing unknown characters.  We cut all sentences to a maximum length of 128 characters. We use a embedding size of 64 for the characters, a single layer LSTM with 2048 Units and a softmax layer implemented with DyNet \cite{dynet} as our neural model. Training is performed with the whole data using Adam \cite{kingma2014adam} by randomly picking a batch until convergence on the validation data. We retrain the resulting model on the Switchboard training data using Stochastic Gradient Descent with a low learning rate of 0.01, which is inspired by \cite{wu2016google}. 
This procedure results in an average entropy of 1.34 bits per character (BPC) on the train set, 1.37 BPC on the validation set and 1.46 BPC on the evaluation set (LDC2002S09).

\subsection{Attention Based Sequence to Sequence Model}

The attention based model was trained on the Switchboard training data. We decode the acoustic model without any linguistic information by applying the greedy method of Section~4.1. The sequence of generated characters is used as the input to our model and we use the sequence of words in the reference transcription as our desired output during training.

For implementing the attention based encoder decoder, we use the Nematus toolkit \cite{nematus}. In our experiments we use GRU units in the encoder and the target sequence is generated using conditional GRU units \cite{nematus}.

We use the default network architecture, with an embedding size of 500 and a hidden layer size of 1024 and our output vocabulary consists of almost 30,000 words. For regularization, we use dropout \cite{hinton2012improving}. Due to time constraints, we only use segments with a maximum length of 100 tokens. The system was trained using Adadelta \cite{zeiler2012adadelta} and a mini-batch size of 80. We performed early stopping on the validation data.

\section{Experiments}

We use the 2000 HUB5 ``Eval2000'' (LDC2002S09) dataset for evaluation. It consist of a ``Switchboard'' subset, which is similar to the training data, and the ``Callhome'' subset. These subsets allow to analyze the robustness of the individual approaches to some extend.

\begin{table}[th]
  \caption{Comparison of Word Error Rates for different decoding approaches on the Eval2000 (E2), Call Home (CH) and Switchboard (SW) (sub-)sets.}
  \label{tab:results}
  \centering
  \begin{tabular}{ c  c  c  c  c}
  Search Method & Ac. Model & E2 & CH & SW \\
  \hline
  \hline
  Greedy & Character  & 37.2\% & 44.0\% & 30.4\% \\
  Char Beam & Character  & 25.1\% & 31.6\% & 18.6\% \\
  WFST & Character  & 23.6\% & 30.2\% & 17.0\% \\
  WFST & Phoneme  & 19.6\% & 25.5\% & 13.6\% \\
  Seq2Seq & Character  & 34.4\% & 40.6\% & 28.1\% \\
  Seq2Seq & Phoneme  & 26.5\% & 33.1\% & 19.8\% \\
  \hline
  Char Beam \cite{maas2015lexicon} & Character  & 30.8\% & 40.2\% & 21.4\% \\ 
  Char Beam \cite{zweig2016advances}  & Character  &  & 32.1\% & 19.8\% \\
  WFST \cite{zweig2016advances} & Character  &  & 26.3\% & 15.1\% \\
  Seq2Seq \cite{zweig2016advances} & Character  &  & 37.1\% & 24.7\% \\
\end{tabular}
\end{table}

For the Greedy Search, we use an alphabet consisting of upper and lowercase characters. As in \cite{zweig2016advances}, an upper case character denotes the start of a new word. For all other character based AMs we only use lowercase characters without a space unit. Table~\ref{tab:results} shows the results, and compares
our findings (top part) against related results from the literature (bottom part).

While Open Vocabulary approaches such as the character RNN approach are still slightly inferior to word-based approaches, adding linguistic information at the character level yields competitive results compared to a tuned word based WFST system. Using the character LM during the Beam Search significantly reduces incorrectly recognized words, which did not appear in the training text (199), by a factor of 30 compared to a simple beam search (6274). This amounts to a rate of 0.5\%, which compares favorably to the out of vocabulary rate of the WFST based approach 0.9\%.
These remaining errors are for the most part very similar to valid English words, and could be considered spelling mistakes (``penately'') or newly created words (``discussly''). Only on rare occasion does the Character LM not add a space between words (``andboxes''). Most notably, the Open Vocabulary approach was able to generate correct words, which did not appear in the training corpora, including ``boger'', ``disproven'', ``ducting'', ``fringing'', ``spick'' and ``peppier''.

\begin{table}[th]
  \caption{Insertion Rate (I), Substitution Rate (S) and Deletion Rate (D) for multiple decoding algorithm using Character based AMs evaluated on the Eval2000 dataset.}
  \label{tab:errors}
  \centering 
  \begin{tabular}{ c  c  c  c }
  Method& I & S & D \\
  \hline
  Greedy & 2.4\% & 26.2\% & 8.6\% \\
  Character Beam & 3.6\% & 16.5\% & 5.0\%\\
  WFST & 8.8\% & 13.0\% & 1.9\% \\
  Seq2Seq & 6.6\% & 21.9\% & 5.9\% \\
\end{tabular}
\end{table}

Table~\ref{tab:errors} shows the insertion, deletion, and substitution rates. We consistently used an insertion bonus of 2.5 in our experiments with the beam search. The application of an insertion bonus every time when reducing the probability by the character based LM yields balanced insertion and deletion errors. Additionally the logarithm of the insertion bonus corresponds well with the entropy of the character language model on the validation set ($log_2(2.5) = 1.3$, validation entropy: 1.37 BPC). Overall, the error patterns of all three systems seem remarkably similar, even though the WFST system has been tuned more aggressively than the other two systems, and thus exhibits unbalanced insertions and deletions.

\begin{table}[th]
  \caption{Example output of a cherry picked utterance}
  \label{tab:example}
  \centering
  \begin{tabular}{ c | c}
  Method & Text \\
  \hline	
  Reference & he is a police officer\\
  Greedy & he'sapolifefolvisere\\
  Character Beam & he's a police officer\\
\end{tabular}
\end{table}

While the attention based system was able to improve over the greedy search results, it did not achieve large gains on the character based AM. We speculate that it might not be the best option to use the transcription of the greedy decoding as the input to the attention based system. We argue that by providing the complete probability distribution over each character the WFST is able to use more information and can outperform the attention based system. Table~\ref{tab:example} shows an example utterance to visualize the characteristics of the different systems.

Our character based system is competitive to the recently published results in \cite{zweig2016advances}, which represent state of the art results. We are within 2\% WER of the reported number for word based WFST. For open vocabulary, character based speech recognition we report an improvement of over 1\% WER compared to previous results \cite{zweig2016advances, maas2015lexicon}.

\section{Conclusions}

In this paper, we compare different decoding approaches for CTC acoustic models, which are trained on the same, open source platform. A ``traditional'' context independent WFST approach performs best, but the open vocabulary character RNN approach performs only about 10\% relative worse, and produces a surprisingly small number of ``OOV'' errors. The Seq2Seq approach produces reasonable performance as well, and is very easy to train, given that the CTC model already produces a symbolic tokenization of the input. This trick allows us to outperform true end-to-end approaches such as \cite{7472641}. 

We believe that these results show that there is currently a multitude of different algorithms that can be used to perform speech recognition in a neural setting, and there may not be a ``one size fits all'' approach for the foreseeable future. While WFST is well understood and fast to execute, the Seq2Seq approach might integrate well with machine translation, while the character RNN approach might perform well for morphologically complex languages. 

We are continuing to further develop these approaches, in order to better understand their characteristics also on non-English data, with different vocabulary growth, or under variable acoustic conditions. 

\section{Acknowledgments}
This work was supported by the Carl-Zeiss-Stiftung and by the CLICS exchange program. 
This work used the Extreme Science and Engineering Discovery Environment (XSEDE), which is supported by National Science Foundation grant ACI-1548562.

\bibliographystyle{IEEEtran}

\bibliography{mybib}

\end{document}